\documentclass[runningheads]{llncs}

\usepackage{eccv}

\usepackage{eccvabbrv}

\usepackage{graphicx}
\usepackage{booktabs}
\usepackage{tabularx}
\usepackage{tabularray}
\usepackage{colortbl}
\usepackage{wrapfig}
\usepackage{algorithm}
\usepackage{algpseudocode}

\usepackage{listings}
\usepackage{xcolor}

\definecolor{codegreen}{rgb}{0,0.6,0}
\definecolor{codegray}{rgb}{0.5,0.5,0.5}
\definecolor{codepurple}{rgb}{0.58,0,0.82}
\definecolor{backcolour}{rgb}{0.95,0.95,0.92}

\lstdefinestyle{mystyle}{
    backgroundcolor=\color{backcolour},   
    commentstyle=\color{codegreen},
    keywordstyle=\color{magenta},
    numberstyle=\tiny\color{codegray},
    stringstyle=\color{codepurple},
    basicstyle=\ttfamily\footnotesize,
    breakatwhitespace=false,         
    breaklines=true,                 
    captionpos=b,                    
    keepspaces=true,                 
    numbers=left,                    
    numbersep=5pt,                  
    showspaces=false,                
    showstringspaces=false,
    showtabs=false,                  
    tabsize=2
}
\usepackage[accsupp]{axessibility}  

\usepackage{hyperref}

\usepackage{orcidlink}

\begin{document}

\title{Self-supervised pretraining for an iterative image size agnostic vision transformer} 

\titlerunning{Self-supervised pretraining for iterative vision transformer}


\author{
Nedyalko Prisadnikov\inst{1} \and
Danda Pani Paudel\inst{1} \and
Yuqian Fu\inst{1} \and
Luc Van Gool\inst{1}
}

\authorrunning{N.~Prisadnikov et al.}

\institute{
$^1$ INSAIT, Sofia University "St. Kliment Ohridski", Bulgaria
}

\maketitle

\begin{center}
\vspace{-0.1in}
\centering
\captionsetup{type=figure}
\includegraphics[width=\textwidth]{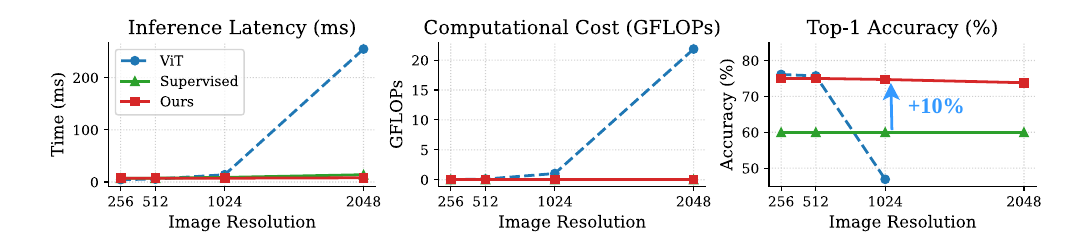}
\vspace{-18pt}
\captionof{figure}{\label{fig:teaser}
\textbf{Computational Efficiency and Scaling.} 
(Left \& Middle) Latency and GFLOPs of a standard ViT versus our foveal model
(evaluated sequentially over 8 steps with a learned gaze policy).
While standard ViT compute scales quadratically with the number of patches,
our dynamic model maintains a strictly $\mathcal{O}(1)$ footprint. 
(Right) ImageNet-1K Top-1 accuracy across extreme evaluation resolutions.
All models are pretrained using an effective global crop size of $256 \times 256$.
While standard ViTs collapse at ultra-high resolutions ($>$1024) due to patch distribution shift,
our image-size agnostic approach maintains robust performance,
outperforming the supervised baseline~\cite{prisadnikov2025vision} by $+10\%$.
}
\end{center}

\begin{abstract}

Vision Transformers (ViTs) dominate self-supervised learning (SSL). 
While they have proven highly effective for large-scale pretraining,
they are computationally inefficient and scale poorly with image size. 
Consequently, foundational models like DINO are constrained to low-resolution processing. 
A recent foveal-inspired transformer~\cite{prisadnikov2025vision}
achieves resolution agnosticism by iteratively processing a fixed-size context of multi-zoom patches. 
This model demonstrated promising results via supervised learning,
utilizing a sequential, recurrent-like process without backpropagation through time. 
To unlock its potential as a foundational backbone,
we introduce a novel \emph{sequential-to-global} SSL framework based on DINO's self-distillation objective. 
Supported by an efficient integral-image patch extraction method,
our approach enables large-scale pretraining for image-size agnostic vision encoders. 
We achieve competitive performance on ImageNet-1K and downstream classification tasks,
maintaining a constant computational budget regardless of input resolution.

\end{abstract}

\section{Introduction}

Self-supervised learning (SSL) serves as the backbone of modern foundational models,
driving breakthroughs in both computer vision~
\cite{caron2021emerging,he2022masked,chen2020simple,he2020momentum,oquab2023dinov2,simeoni2025dinov3}
and natural language processing (NLP)~
\cite{devlin2019bert,brown2020language,mikolov2013efficient,pennington2014glove,radford2018improving,liu2019roberta,raffel2020exploring}.
By alleviating the burden of large-scale manual annotation,
SSL enables pretraining at an unprecedented scale.
Within computer vision, this progress has been largely fueled through the Vision Transformer
(ViT~\cite{dosovitskiy2020image}) architecture.
ViT-based encoders have become the standard for state-of-the-art vision models~
\cite{caron2021emerging,oquab2023dinov2,simeoni2025dinov3}
and the visual components of Multimodal Large Language Models
(MLLMs)~\cite{radford2021learning,zhai2023sigmoid,tschannen2025siglip, fu2025objectrelator}.

Despite this success,
the standard ViT architecture presents significant challenges
regarding efficiency.
Modern visual streams -- from high-resolution cameras to robotic sensors --
contain tens or hundreds of megapixels.
Yet, foundational models like DINO are largely restricted to pretraining on small resolutions
(typically $224 \times 224$).
This limitation stems from the architecture's computational complexity,
which scales poorly with the input image size.

To address this,
recent work~\cite{prisadnikov2025vision} argues that vision encoders
should be image-size agnostic and task-driven.
It proposed a dynamic, active vision transformer inspired by the foveal structure of the human visual system.
Instead of processing a full grid of patches at once,
the model operates iteratively.
At each step, it processes a small, fixed number of ``multi-zoom'' patches around a chosen gaze location,
maintaining a set of learnable state tokens that act as an evolving internal memory between steps.
Crucially, these multi-zoom patches are extracted in a top-down manner:
their dimensions are defined as a ratio of the overall image size rather than a fixed absolute pixel count.
Because standard ViTs extract fixed-size patches (e.g. $16 \times 16$ pixels),
increasing input resolution alters the semantic content within a single patch,
leading to a distribution shift~\cite{touvron2019fixing} and performance collapse
(Fig.~\ref{fig:teaser}, right).
By utilizing ratio-based patches, the foveal approach eliminates this distribution shift.
A reinforcement learning policy then dictates the sequence of gaze locations,
creating a recurrent-like system that accumulates scene understanding without
backpropagation through time (BPTT)~\cite{werbos1990backpropagation}.

While this architecture is promising, its initial validation was limited to
supervised learning on ImageNet-1k~\cite{deng2009imagenet}.
For such a dynamic foveal architecture to serve as a true foundational backbone,
it must be compatible with self-supervised pretraining.
However, adapting state-of-the-art SSL frameworks like DINO~\cite{caron2021emerging}
to a recurrent-like, stop-gradient architecture is highly non-trivial
due to the sequential nature of the model.
Furthermore, the patch extraction mechanism in~\cite{prisadnikov2025vision}
scales computationally with image resolution.
This creates a bottleneck that prevents true image-size agnosticism,
particularly under the high-throughput demands of training.

In this paper, we resolve both of these critical limitations.
We demonstrate, for the first time, that a dynamic foveal transformer can be pretrained
in a self-supervised manner without BPTT.
We achieve this by radically adapting the DINO self-distillation objective.
Standard DINO utilizes a spatial multi-crop setting~\cite{caron2020unsupervised},
matching local spatial crops to global spatial views.
We formulate a novel \emph{sequential-to-global} distillation process.
We demonstrate that a sequence of foveal glimpses
can be treated analogously to the standard random resized local crops of DINO,
training the sequentially evolving state tokens to match the embeddings of a standard,
global ViT teacher.
This hybrid formulation teaches the model to dynamically assemble a global understanding
of the scene from iterative, local multi-zoom views.

The main \underline{contributions} of this work are as follows:
\begin{itemize}
\item We propose a \textbf{sequential-to-global self-supervised pretraining framework}
for dynamic vision transformers.
By mapping sequences of local foveal glimpses to global spatial views without BPTT,
we achieve a model capable of evolving its scene representation
iteratively while leveraging DINO's self-distillation objective.
\item We extend the multi-zoom patch mechanism from~\cite{prisadnikov2025vision}
by introducing a \textbf{dense foveal grid} (Sec.~\ref{sec:foveal-extract}) around the center of gaze.
This provides a richer, high-resolution context for the self-attention mechanism.
\item We introduce an efficient \textbf{patch extraction method}
(Sec.~\ref{sec:foveal-extract}) based on integral images.
By isolating the resolution-dependent computation to a single,
one-time preprocessing step,
the subsequent iterative extraction of foveal patches becomes constant with respect to image size.
See comparison with~\cite{prisadnikov2025vision} in Fig.~\ref{fig:patchify-mflops}.
\end{itemize}

We validate our learned representations on ImageNet-1K~\cite{deng2009imagenet} through linear probing.
Operating as a standard ViT, our ViT-Small equivalent achieves 76\% top-1 accuracy,
competitive with DINO~\cite{caron2021emerging}.
Crucially, in its dynamic 8-step foveal mode,
a learned gaze policy yields 75\% accuracy --
significantly outperforming random gazes (71.7\%).

Although standard classification limits the spatial search advantages of an active policy,
these results prove our model can build accurate global representations from sparse, localized glimpses.
Most importantly, it achieves this with a fixed per-step computational budget decoupled from input resolution.
Consequently, our image-size agnostic approach avoids the standard ViT performance collapse
when evaluated on significantly larger images compared to training (Fig.~\ref{fig:teaser}).

\section{Preliminaries}

As outlined in the introduction,
our method builds upon the dynamic,
fovea-inspired vision transformer proposed by Prisadnikov et al.~\cite{prisadnikov2025vision}.
To keep this paper self-contained,
this section briefly reviews the core architectural setting introduced in that work.
Subsequently, in Sec.~\ref{sec:method},
we present our contributions and detail the proposed self-supervised pretraining framework.

\subsection{Overall Architecture}

The dynamic foveal encoder from~\cite{prisadnikov2025vision} consists of three main components: 
1) a \textbf{Transformer} backbone that serves as the primary visual processor; 
2) a \textbf{Task Head} that decodes the transformer's latent state into task-specific predictions
(e.g., classification logits); and 
3) a \textbf{Gaze Policy} that determines the fixation point $(x, y)$ for the subsequent iteration,
effectively dictating \emph{where} to look next.

Architecturally, the main processor is identical to a standard Vision Transformer (ViT),
with the primary distinction being the inclusion of multiple \texttt{CLS} tokens.
These multiple \texttt{CLS} tokens represent the evolving latent state,
acting as the model's working memory of the scene explored thus far.
As this working memory is a bottleneck,
using multiple state tokens increases its representational capacity.
It shares similarities with the object queries used in DETR~\cite{carion2020end}
and the register tokens introduced in~\cite{darcet2023vision}.

The processing unfolds iteratively.
At the initial step ($t=1$), the state tokens $s_1$ are initialized as learned embeddings.
The policy network takes $s_1$ as input and predicts the first gaze location.
Following this, a set of multi-zoom patches $p_1$
(Sec.~\ref{sec:mz-patches}) is extracted around this gaze location.
The transformer then processes the sequence of concatenated tokens $[s_1, p_1]$.
The output corresponding to the input state tokens becomes the updated state $s_2$ for the next step.
At this point, the task head can process $s_2$ to yield the final prediction,
or the policy can process $s_2$ to select a new gaze location,
extract new patches $p_2$,
and run the transformer again to produce $s_3$.
This active exploratory process continues as long as necessary for the task,
or for a predefined number of steps.

It is important to note that
while the sequence of state tokens $s_t$ is reminiscent of a hidden state in a Recurrent Neural Network (RNN),
\emph{no Backpropagation Through Time} is applied.
Before $s_t$ is fed into the $t$-th iteration of the transformer,
it is detached from the computation graph.
Consequently, no gradients flow across different temporal steps,
ensuring training remains efficient.

\subsection{Gaze Policy}

The gaze policy is the defining component
that separates this active architecture from
standard, passive ViT encoders.
Given the current state tokens $s_t$,
the goal of the policy is to predict the continuous coordinates
$(x, y)$ of the next gaze fixation.
Because the patch extraction process is non-differentiable,
the policy is trained separately from the main transformer via Reinforcement Learning (RL).
Both in our work and in~\cite{prisadnikov2025vision},
the policy is trained using policy gradient optimization~\cite{williams1992simple},
specifically leveraging a variant of Group Relative Policy Optimization (GRPO)~\cite{shao2024deepseekmath}.

\subsection{Multi-zoom Top-Down Patches}
\label{sec:mz-patches}

In order to be image size agnostic,
the input context to the transformer must remain fixed in length;
i.e., the number of patches fed to the transformer at every step must be independent of the input image resolution.
Furthermore, these patches must be extracted in a \emph{top-down} manner,
meaning that the crop size representing a patch is defined as a relative ratio of the image dimensions. 

This contrasts sharply with the bottom-up approach of standard ViTs,
which extracts patches based on a fixed pixel grid
(e.g., $16 \times 16$ pixels).
As discussed in the introduction,
the bottom-up approach suffers from distribution shifts:
if the image resolution is increased post-training,
a $16 \times 16$ patch captures a significantly smaller semantic region,
forcing the tokenizer to process unfamiliar levels of detail~\cite{touvron2019fixing}.
As demonstrated in Fig.~\ref{fig:teaser}
this can lead to performance collapse when the size disparity is large.

In the framework of~\cite{prisadnikov2025vision},
a patch can be any square crop of the image.
Regardless of its original crop size,
the region is simply resized to a fixed dimension
($16 \times 16$ pixels) before being processed by the \texttt{PatchEmbed} tokenizer.
This mechanism resembles biological vision.
A patch derived from a small crop will maintain high resolution
(retaining fine details after resizing) but will have narrow spatial coverage,
akin to the $1^{\circ}\text{-}2^{\circ}$ central angle of the human fovea.
Conversely, a patch derived from a large crop will cover a broad region of the image
but lose high-frequency details during resizing,
similarly to our low-acuity peripheral vision.

\begin{figure}[tb]
    \centering
    \begin{minipage}{0.4\textwidth}
        \centering
        \includegraphics[width=\linewidth]{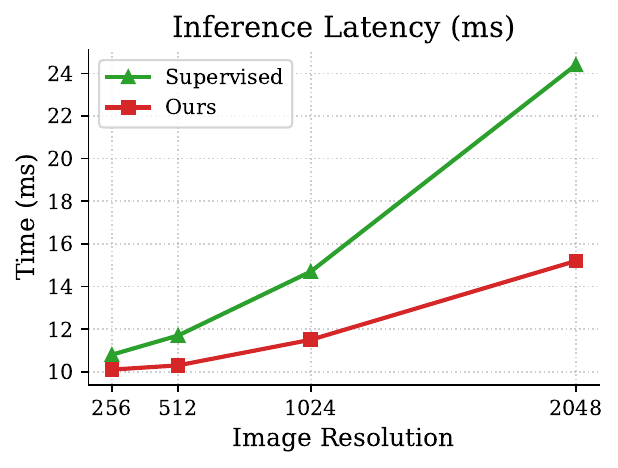}
    \end{minipage}\hfill
    \begin{minipage}{0.5\textwidth} 
\caption{\textbf{Foveal Extraction Latency.} 
Median latency for extracting 1,000 patches (10 steps of 100 patches)
from a single image, simulating a high-throughput training workload. 
Including running the models.
Compared to the supervised baseline~\cite{prisadnikov2025vision},
our integral image approach reduces overhead. 
The increase of latency with image size reflects the one-time cost for converting the image into a integral one.
}
        \label{fig:patchify-mflops}
    \end{minipage}
\end{figure}

Specifically, the multi-zoom patch mechanism extracts
a sequence of $M$ concentric patches sharing the same gaze center $(x, y)$,
with crop sizes growing exponentially.
The side length of each square crop, $S_{\text{patch}}$, is defined as:
\begin{equation}
S_{\text{patch}} = \frac{\min(H, W)}{2^z}, \quad z \in \mathrm{linspace}(0.0, 5.0, M)
\end{equation}
where $H$ and $W$ denote the image height and width,
and $z$ represents the zoom level.
Examples of these extracted multi-zoom patches are illustrated in Fig.~\ref{fig:mz-focused-patches}.
\section{Method}
\label{sec:method}

This section details our framework for training a dynamic active vision transformer. 
We first introduce our novel sequential-to-global self-supervised pretraining objective (Sec.~\ref{sec:dino}). 
We then describe the mechanisms that enable resolution-agnostic,
$\mathcal{O}(1)$ foveal extraction (Sec.~\ref{sec:foveal-extract}), 
and finally outline the reinforcement learning formulation used to train the active gaze policy (Sec.~\ref{sec:grpo}).

\subsection{Sequential-to-Global Self-Supervised Pretraining}
\label{sec:dino}

\begin{figure}[tb]
    \centering
    \includegraphics[width=\textwidth]{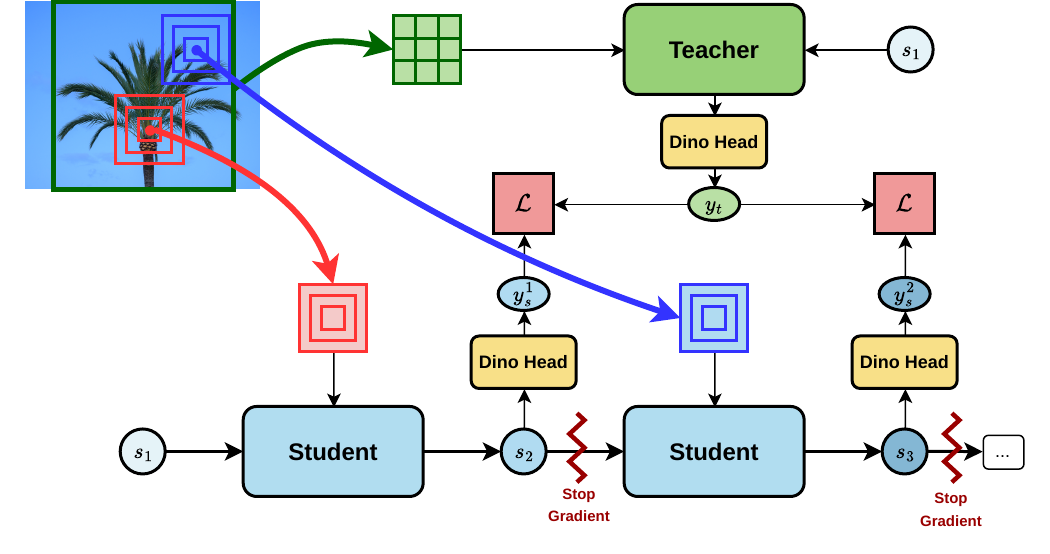}
    \caption{Main architecture for sequential-to-global pretraining.
    The student network iteratively updates its state tokens,
    computing the distillation loss at each step against a static teacher embedding.
    Stop-gradients prevent backpropagation through time.}
    \label{fig:main_arch}
    \vspace{-0.1in}
\end{figure}

Our pretraining approach builds upon the self-distillation framework
of DINO~\cite{caron2021emerging,oquab2023dinov2,simeoni2025dinov3}.
Standard DINO trains a student network to match a momentum-updated teacher network
by comparing different spatial crops of the same image,
typically teaching local-to-global correspondence.
We fundamentally extend this paradigm to a \emph{sequential-to-global} framework,
training an iterative foveal model to accumulate scene understanding over time
without backpropagation through time (BPTT).

\textbf{Implicit Cropping via Top-Down Grids.} 
Standard ViTs crop and resize the image before dividing it into bottom-up patches.
To maintain image-size agnosticism,
we replace cropping with implicit top-down grid sampling.
To simulate a DINOv2 global crop
(i.e., a $224 \times 224$ pixel area divided into $14 \times 14$ pixel patches,
yielding a spatial grid of $16 \times 16$ patches),
we simply center a foveal grid (Sec.~\ref{sec:foveal-extract}) at the crop's intended location
on the original, unresized image.
The zoom level is adjusted so the foveal grid mathematically covers the exact same area ratio as the target crop.
This allows us to extract standard DINO views efficiently in a top-down manner.
Refer to the supplementary material for the exact area-to-zoom mathematical mapping.

\textbf{Sequential-to-Global Distillation.} 
Instead of relying strictly on independent spatial augmentations,
we introduce an iterative, step-wise sequence of views.
We define three distinct view types for a single image:
\begin{enumerate}
\item \textbf{Global view:} A spatial grid of $16 \times 16$ foveal patches covering $>0.32$ of the image area.
\item \textbf{Local view:} A spatial grid of $7 \times 7$ foveal patches covering $<0.32$ of the image area.
\item \textbf{Sequential view:}
A dynamic sequence of 8 random foveal glimpses processed iteratively, with state tokens passed between steps.
\end{enumerate}

As in DINO, the networks receive asymmetric views.
The \textbf{teacher} processes one global view and one sequential view,
but only generates target embeddings from the \emph{final} step of the sequence. 
The \textbf{student} processes a different global view,
8 independent local views, and a different sequential view. 
Crucially, for the sequential view,
the student considers the output at \emph{every} step.
As illustrated in Fig.~\ref{fig:main_arch},
after each foveal glimpse, the student's current state is projected to match the teacher's static targets.
Between steps, gradients are stopped.
This forces the student to iteratively approximate the global context from local multi-zoom patches
while preventing the computational explosion of BPTT.
We compute the loss by matching all student outputs to all teacher outputs.
Note that Fig.~\ref{fig:main_arch} shows only a single teacher-student combination for simplicity.

\textbf{Objective and Projection Head.} 
We optimize the standard cross-entropy loss $\mathcal{L} = -\sum p_t \log p_s$,
utilizing the Sinkhorn-Knopp online clustering from DINOv2~\cite{oquab2023dinov2}.
Standard DINO models project a single \texttt{[CLS]} token through a multi-layer perceptron (MLP) head.
Because our dynamic transformer maintains multiple state tokens ($N_{state}$),
we apply the projection MLP exclusively to the first state token.
This design explicitly bottlenecks the high-level semantic understanding
required for the DINO objective into this primary token,
leaving the remaining tokens to function freely as unconstrained internal memory between glimpses.

The result of this training process is a \emph{hybrid} transformer
capable of operating natively as a standard ViT (using a single global top-down grid)
or as an active foveal transformer with iterative memory.
In the supplementary material, please find a simplified pseudo-code
of the training loop.

\subsection{Efficient Foveal Extraction}
\label{sec:foveal-extract}

To enable the sequential processing described above, the model requires a context that captures both broad peripheral understanding and fine-grained central details, without scaling computationally with the image resolution.

\begin{figure}
\centering
\vspace{-0.3in}
\includegraphics[width=0.9\linewidth]{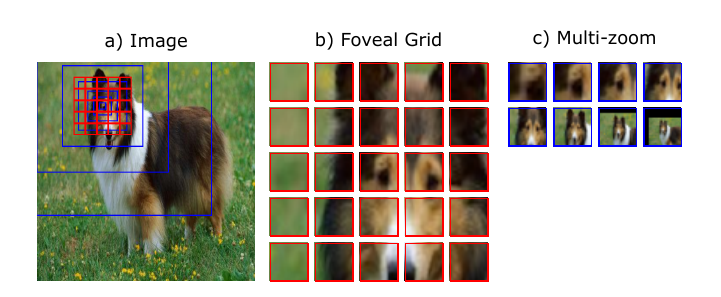}
\caption{
Combining multi-zoom patches with focused grid of patches for informative context for the transformer.
a) Shows the input image with the bounding-boxes of the multi-zoom crops (blue)
and the foveal grid crops (red) overlayed, b) shows the extracted grid patches, and
c) shows the multi-zoom patches.
Note that the patches are allowed to span outside the image area
by filling them with zero padding.
}
\label{fig:mz-focused-patches}
\vspace{-0.2in}
\end{figure}

\noindent \textbf{Augmenting Multi-Zoom Context with a Foveal Grid.} 
While the logarithmic multi-zoom patches from~\cite{prisadnikov2025vision}
provide excellent peripheral context,
they yield a sparse representation of the high-resolution center. 
To provide sufficient granular context for the self-attention mechanism,
we introduce a \emph{Focused Foveal Grid}:
a dense $G \times G$ grid of patches extracted around the gaze center. 
Since $G$ is a fixed constant, the total sequence length of the composite context
(multi-zoom + foveal grid) remains strictly constant
regardless of input resolution.
Unlike standard bottom-up ViT patches where the token count scales with image area, our grid is \emph{top-down}. 
The side length of each patch is defined relative to the image dimensions ($S_{patch} = \min(H, W) / 2^z$, where $z$ is the zoom level). 
The zoom level $z$ is selected from a predefined zoom range
which determines the spatial coverage of the foveal grid (independently from the image size).
See an example foveal context in Fig.~\ref{fig:mz-focused-patches}.
The details of the top-down foveal grid formulation is presented in the supplementary material.

\noindent \textbf{Resolution-Agnostic Extraction via Integral Images.} 
Although top-down patching fixes the token count,
standard patch extraction operations still exhibit linear complexity $\mathcal{O}(H_c \times W_c)$ with respect to the crop's pixel count.
To eliminate this bottleneck,
we utilize an \emph{Integral Image} (Summed-Area Table)~\cite{ViolaJ01, Crow1984SummedareaTF} representation. 
By pre-computing the integral image in a single pass,
we can extract and resize any rectangular patch using standard box-filtering. 
This reduces the amortized cost of iterative patch extraction to $\mathcal{O}(1)$ time complexity per target pixel,
decoupling our dynamic foveal operations from the source image size (Fig.~\ref{fig:patchify-mflops}).
Note that the conversion of the image to an integral one is performed only once per image and then we can efficiently extract many patches.
More details on extracting patches from integral images can be found in the supplementary material.

\subsection{Gaze Policy Optimization via GRPO}
\label{sec:grpo}

The final component of our framework is the active gaze policy, which directs the frozen transformer where to look based on the evolving internal state.

\noindent \textbf{Policy Distribution.} 
Because the composite foveal context is defined only the gaze center,
the policy only needs to predict the relative $(x, y)$ coordinates for the next fixation. 
As optimal gaze locations are often highly multi-modal
(e.g., multiple distinct object parts),
a single Gaussian is insufficient. 
We formulate the policy as a continuous Gaussian Mixture Model (GMM) with a fixed diagonal variance schedule. 
Architecturally, the policy is parameterized by a lightweight transformer
that processes the $N_{state}$ memory tokens alongside learnable query tokens,
which predict the spatial means and mixture weights for each GMM component.

\noindent \textbf{Optimization Objective.} 
We optimize the policy using a variant of Group Relative Policy Optimization (GRPO) featuring per-image and per-step return normalization. 
For a given image, we collect a group of $G$ traces, where each trace $o_i$ consists of $T$ sequential steps. 
The immediate reward $R_{i,t}$ at step $t$ is defined as the predicted probability of the ground-truth class $y$ given the task head $\tau$:
\begin{equation}
R_{i,t} = \mathbb{P}(\tau(s_{t+1}) = y).
\end{equation}
We compute discounted returns $G_{i,t} = \sum_{t'=t}^{T}{\gamma^{t'-t}R_{i,t}}$ and calculate group-normalized advantages $\hat{A}_{i,t} = (G_{i,t} - \bar{G}_t) / \hat\sigma(G_t)$. 
By rewarding the policy based on the classification confidence at each step, the model learns to actively seek out discriminative spatial regions that maximize the frozen backbone's understanding of the scene.

In essence our approach to the gaze policy optimization mimicks the one from~\cite{prisadnikov2025vision}
besides using a different immediate reward.
We found that this reward yields slightly better performance than the ones used in~\cite{prisadnikov2025vision}.

\section{Architectures}
\label{sec:arch}

\textbf{Main Transformer.}
We adopt the ViT-Small architecture~\cite{dosovitskiy2020image}
following the DINOv2~\cite{oquab2023dinov2} implementation,
modified for active foveal vision as proposed in~\cite{prisadnikov2025vision}. 
The network consists of 12 blocks with an embedding dimension of 384 distributed across 6 attention heads. 
We maintain $N_{state} = 8$ internal state tokens
and use a patch size of 16x16 pixels
(i.e., each extracted foveal crop is resized to a 16x16 spatial grid before tokenization).

\noindent \textbf{DINO Projection Head.}
Following DINOv2~\cite{oquab2023dinov2},
we apply a 3-layer MLP projection head exclusively to the \emph{first} state token. 
This structural bottleneck explicitly forces the first state token
to accumulate a high-level semantic understanding of the scene. 
The head projects the token into a 65,536-dimensional space
to match the cluster prototypes in the Sinkhorn-Knopp algorithm. 
(Note: Attempts to pool all state tokens using a transformer-based head led to representational collapse,
likely because an identical meta \texttt{[CLS]} token across both teacher and student
bypasses the need to align the actual state tokens).

\noindent \textbf{Task Head.}
Because we evaluate our representations via linear probing and classification,
the task head maps the frozen network's output to dataset-specific class logits. 
We evaluate two designs:
a linear layer applied strictly to the first state token,
and a single-layer transformer that pools information across all $N_{state}$ tokens before projection. 
Empirically, the simpler linear head performs best on ImageNet-1K~\cite{deng2009imagenet},
while the transformer head excels on fine-grained datasets requiring accumulated details,
such as CUB-200-2011~\cite{wah2011caltech} and Oxford 102 Flowers~\cite{Nilsback08flowers}.

\noindent \textbf{Gaze Policy.}
We parameterize the active vision policy as
a 6-block transformer that outputs the parameters of a Gaussian Mixture Model (GMM). 
To predict the parameters for $N_{mixture}$ components,
the network processes the $N_{state}$ state tokens
alongside $N_{mixture} + 1$ learnable query tokens. 
The first $N_{mixture}$ queries predict the spatial means of the Gaussian components,
while the final query outputs the categorical logits for the mixture weights.
\section{Experiments}

\begin{table}[tb]
\centering
\caption{\textbf{Main Results (Top-1 Accuracy).}
Evaluation of our frozen sequential-to-global representations via linear probing
on ImageNet-1K, CUB, and Flowers.
We compare the model operating as a dense, static ViT against its dynamic foveal modes
(Random Gazes vs. Learned Policy).
The dynamic mode results are reported after 8 iterative steps.
}
\label{tab:results}
\begin{sc}
\begin{tabular*}{\textwidth}{@{\extracolsep{\fill}} l ccc ccc ccc}
\toprule
& \multicolumn{3}{c}{ImageNet} & \multicolumn{3}{c}{CUB} & \multicolumn{3}{c}{Flowers}\\
\cmidrule(lr){2-4} \cmidrule(lr){5-7} \cmidrule(lr){8-10}
Variant & ViT & Rand & Policy & ViT & Rand & Policy & ViT & Rand & Policy\\
\midrule
Ours & 76.1 & 71.7 & 75.0 & \textbf{81.1} & 73.8 & 80.5 & \textbf{98.3} & 97.6 & 97.7 \\
DINO & \textbf{77} & & & 78.6$^\dagger$ & & & 96.4 & & \\
Supervised~\cite{prisadnikov2025vision} & & 60.0 & 65.0 & & & & & & \\
\bottomrule
\end{tabular*}
\end{sc}
\vspace{2pt}
\raggedright \footnotesize \textnormal{$^\dagger$ Baseline performance reported by~\cite{chowdhury2025prompt}.}
\end{table}
We evaluate our sequential-to-global framework primarily on ImageNet-1K~\cite{deng2009imagenet},
a standard benchmark for validating self-supervised representations.
To assess the generalizability and rich contextual detail of our learned representations,
we further evaluate the frozen model on fine-grained visual classification (FGVC) benchmarks:
CUB-200-2011~\cite{wah2011caltech} and Oxford 102 Flowers~\cite{Nilsback08flowers}.

\subsection{Implementation Details}
Our pipeline consists of three distinct modules:
the main foveal transformer,
the task-specific classification head,
and the active gaze policy.
We train these components in a sequential three-stage pipeline.
During each stage, \emph{all components trained in previous stages are frozen}:
\begin{enumerate}
\item \textbf{Self-Supervised Pretraining:}
The main foveal transformer is pretrained on ImageNet-1K without labels
using our sequential-to-global distillation objective.
\item \textbf{Linear Probing (Task Head):}
The transformer is frozen.
We train a linear classification head specific to the target dataset,
analogous to the standard DINO linear evaluation protocol.
We use random gaze locations to sample input state for the task head.
\item \textbf{Policy Learning:}
Both the transformer and task head are frozen.
We train the task-specific active gaze policy via GRPO to optimize glimpse locations.
\end{enumerate}

\subsection{Pretraining Setup}
We instantiate our main transformer using the ViT-Small architecture.
The model is pretrained on ImageNet-1K with a batch size of 1024.
We use a base learning rate of 0.002,
applying a linear warmup for the first 10 epochs followed by a cosine decay schedule~\cite{loshchilov2016sgdr}.
Following DINO~\cite{caron2021emerging},
the weight decay follows a cosine schedule from 0.04 to 0.4,
and the student temperature $\tau_s$ is fixed at 0.1
while the teacher temperature $\tau_t$ is linearly ramped up from 0.04 to 0.07 over the first 30 epochs.
We replace DINOv1's teacher output centering with Sinkhorn-Knopp~\cite{cuturi2013sinkhorn}
online clustering~\cite{oquab2023dinov2,asano2019self}.
Code to reproduce our pretraining framework and foveal extraction will be made publicly available.

\subsection{Main Results}
For each dataset, we report performance across three evaluation modes using the exact same frozen backbone and task head: 
(1) \textbf{ViT Mode}, serving as the baseline, using a static 16x16 grid of patches covering the full image; 
(2) \textbf{Random Gazes}, processing the image sequentially over 8 steps with randomized locations; and 
(3) \textbf{Learned Policy}, processing the image sequentially guided by the trained gaze module. Results are summarized in Table~\ref{tab:results}.

\textbf{Competitive Baseline Performance.}
When operating in ViT mode, our self-supervised model achieves 76.1\% Top-1 accuracy on ImageNet-1K.
This is highly competitive with standard DINOv1 baselines,
proving that our top-down implicit cropping mechanism successfully captures robust global representations.

\textbf{Efficacy of the Gaze Policy.}
While the dynamic foveal evaluation (75.0\%) closely trails the dense ViT baseline,
the learned gaze policy yields a large improvement over random gazes (71.7\%),
particularly in the early steps.
This demonstrates the policy's ability to actively seek highly discriminative spatial regions.
Most importantly, the foveal model achieves this 75.0\% accuracy with a fixed computational budget per step,
independent of image size.
Paving the way for pretraining on high-resolution images.

\textbf{Monotonic Accumulation of Understanding.}
A critical property of our sequential-to-spatial distillation is that performance
consistently improves monotonically at each step $t$ compared to step $t-1$.
We observed this consistency not just during final evaluation,
but throughout the entire training process.
This validates that the unconstrained state tokens successfully act as an evolving internal memory,
aggregating scene context sequentially without the computational overhead of backpropagation through time.

\begin{table}[tb]
\centering
\begin{small}
\caption{\textbf{Sequential Evaluation on ImageNet-1K.}
Step-by-step Top-1 accuracy on ImageNet in dynamic iterative mode.
The active gaze policy consistently outperforms random glimpses,
and the dynamic model demonstrates monotonic accumulation of scene understanding over time,
remaining competitive with standard DINOv1 baselines.
}
\label{tab:seq-results}
\begin{sc}
\begin{tabular*}{\textwidth}{@{\extracolsep{\fill}} l cccccccc}
\toprule
Variant & \multicolumn{8}{c}{Step}\\
\cmidrule(lr){2-9}
& 1 & 2 & 3 & 4 & 5 & 6 & 7 & 8\\
\midrule
Random Gazes & 45.5 & 57.3 & 62.9 & 66.3 & 68.4 & 70.0 & 70.9 & 71.7\\
w/ Policy & 57.7 & 66.7 & 70.3 & 72.2 & 73.4 & 74.0 & 74.6 & \textbf{75.0}\\
\bottomrule
\end{tabular*}
\end{sc}
\end{small}
\end{table}

\subsection{Ablations}

We ablate the foveal grid size and the number of state tokens using a compute-efficient pretraining setup
(100 epochs, reduced crop resolutions; see Supplementary Material for full details).
To isolate the impact of these components,
we evaluate the exact same frozen model under two linear probing modes:
\textbf{ViT} (standard global processing~\cite{caron2021emerging}) and
\textbf{Foveal} (iterative processing over 8 random gaze locations).
For the Foveal evaluation,
we do not train a policy; 8 random gazes provide sufficient spatial coverage to classify the image.

Table~\ref{tab:ablate-grid-size} demonstrates that the dense foveal grid is critical.
Strictly relying on multi-zoom patches starves the self-attention mechanism of high-resolution context,
significantly degrading accuracy across both evaluation modes.
Furthermore,
Table~\ref{tabl:ablate-n-state} ablates the temporal memory capacity via the number of state tokens ($N_{state}$).
While using multiple state tokens is beneficial for propagating information between glimpses,
adding too many yields diminishing returns. Based on this, we adopt $N_{state}=8$ for our primary experiments.

\begin{table}[tb]
\centering
\begin{minipage}{0.45\textwidth}
\centering
\caption{
\textbf{Foveal grid size ablation}.
 Results show the importance of augmenting the multi-zoom patches
with focused foveal grid of high-resolution patches around the gaze.}
\label{tab:ablate-grid-size}
\setlength{\tabcolsep}{8pt}
\begin{sc}
\begin{tabular}{l cc}
\toprule
& \multicolumn{2}{c}{Top-1 Acc} \\
\cmidrule(lr){2-3}
Grid Size & ViT & Foveal\\
\midrule
1 & 60.7 & 48.0\\
2 & 61.1 & 50.4\\
3 & 63.9 & 55.7\\
4 & 63.7 & 58.3\\
5 & 64.2 & 60.7\\
\bottomrule
\end{tabular}
\end{sc}
\end{minipage}
\begin{minipage}{0.45\textwidth}
\centering
\caption{
\textbf{Number of state tokens ablation}.
Adding multiple state tokens has positive effect on the performance of the model,
but it is not as critical as the foveal grid.
}
\label{tabl:ablate-n-state}
\setlength{\tabcolsep}{8pt}
\begin{sc}
\begin{tabular}{l cc}
\toprule
& \multicolumn{2}{c}{Top-1 Acc}\\
\cmidrule(lr){2-3}
$N_{state}$ & ViT & Foveal\\
\midrule
1 & 62.6 & 59.8\\
4 & 63.7 & 60.6\\
8 & 64.2 & 60.7\\
16 & 62.8 & 59.7\\
\bottomrule
\end{tabular}
\end{sc}
\end{minipage}
\vspace{-0.2in}
\end{table}
\section{Related Work}

\subsection{Self-supervised Learning}

The use of self-supervised learning (SSL)
to eliminate the need for manual labels has been a central goal
since the inception of the modern deep learning era.
Before SSL became a standard term in computer vision,
the Natural Language Processing (NLP) community pioneered the concept of learning from data itself,
leveraging the insight that language contains its own supervision:
the semantic meaning of a word can be inferred from its context.
Foundational works include Word2Vec~\cite{mikolov2013efficient} and GloVe~\cite{pennington2014glove}.
The introduction of the Transformer architecture~\cite{vaswani2017attention}
enabled the modeling of long-range dependencies,
leading to seminal models like
BERT~\cite{devlin2019bert}, which utilizes a Masked Language Modeling (MLM) objective,
and GPT~\cite{radford2018improving}, which employs a causal modeling objective.
These works demonstrated the efficacy of self-supervised training at a massive scale.

Translating SSL to computer vision initially proved challenging due to the high-dimensional,
continuous nature of images compared to discrete text tokens.
Early efforts focused on "pretext tasks" --
auxiliary objectives derived from the image structure -- such as
relative patch prediction~\cite{doersch2015unsupervised},
jigsaw puzzle solving~\cite{noroozi2016unsupervised},
rotation prediction~\cite{gidaris2018unsupervised},
and generative tasks like colorization~\cite{zhang2016colorful} and inpainting~\cite{pathak2016context}.

A major paradigm shift occurred with \emph{contrastive learning},
which moved away from handcrafted pretext tasks toward instance discrimination:
minimizing the distance between augmented views of the same image while maximizing the distance to others.
SimCLR~\cite{chen2020simple} established a simple framework for this
using the InfoNCE loss~\cite{oord2018representation},
while MoCo~\cite{he2020momentum} addressed batch size limitations via a dynamic queue and momentum encoder.
Subsequently, BYOL~\cite{grill2020bootstrap} and SimSiam~\cite{chen2021exploring}
demonstrated that negative pairs were not strictly necessary,
preventing representational collapse through asymmetric predictor networks
and stop-gradient operations.
SwAV~\cite{caron2020unsupervised} introduced online clustering to the contrastive framework.
Building on these, DINO~\cite{caron2021emerging} formulated SSL as self-distillation with no labels,
showing that vision transformers trained this way learn explicit semantic segmentation information.

Most recently,
the field has explored Joint-Embedding Predictive Architectures~\cite{lecun2022path}(JEPA).
Unlike generative approaches such as Masked Autoencoders~\cite{he2022masked}
that reconstruct pixels, methods like I-JEPA~\cite{assran2023self} and LeJEPA~\cite{balestriero2025lejepa}
predict the latent representation of a target block given a context block.
This encourages the model to capture high-level semantic structures
rather than low-level high-frequency details,
effectively learning an internal world model.

\subsection{Iterative Processing and Memory in Transformers}

Prisadnikov et al.~\cite{prisadnikov2025vision} argue that vision encoders
should operate as a sequential process involving small context windows and an evolving memory,
rather than processing full images in a single pass.
This section reviews architectures that integrate recurrence and memory
into the Transformer~\cite{vaswani2017attention} backbone. 
Recurrent Vision Transformers (RVTs)~\cite{gehrig2023recurrent}
introduce a multi-stage backbone for event-based vision,
using LSTMs~\cite{hochreiter1997long} to maintain a hidden state $H_t$ updated with new event features.
In the NLP domain,
the Recurrent Memory Transformer (RMT)~\cite{bulatov2022recurrent}
appends special "memory tokens" to the input sequence.
After a forward pass,
the output embeddings of these tokens are treated as the state and prepended to the input of the next segment,
effectively extending context length indefinitely.
Similarly, Memformer~\cite{wu2022memformer}
utilizes an external dynamic memory module with a read/write mechanism via cross-attention.
Our approach strictly follows the iterative processing proposed in~\cite{prisadnikov2025vision},
which is architecturally similar to RMT but applied to spatial visual exploration.
Crucially, unlike RMT which uses Backpropagation Through Time (BPTT)~\cite{werbos1990backpropagation},
we employ a truncated optimization strategy
where gradients are not propagated between iterations to maintain efficiency.

\subsection{Foveal-inspired vision processing}

The concept of replicating foveal active vision in neural networks
dates back to seminal works like~\cite{schmidhuber1991learning}.
Mnih et al.~\cite{mnih2014recurrent} later introduced the Recurrent Attention Model (RAM),
which processes image data using "glimpses" --
concentric crops similar to the multi-zoom patches in~\cite{prisadnikov2025vision}.
However, these earlier works predated modern components like the Vision Transformer.
More recently,
the Foveated Transformer (FoveaTer)~\cite{jonnalagadda2021foveater}
employed pooling regions to perform classification using a ViT architecture.
The Human Attention Transformer (HAT)~\cite{yang2024unifying}
integrates visual information at two eccentricities (foveal and peripheral) to predict human scanpaths --
the temporal sequence of fixations and saccades~\cite{noton1971scanpaths}.
This architecture mimics the biological
"dual-stream" processing hypothesis~\cite{goodale1992separate},
where the ventral stream handles object identification (high-resolution foveal)
and the dorsal stream handles spatial location (low-resolution peripheral).
Also relevant is Fovea-Like Input Patching (FLIP)~\cite{traub2025looking},
which achieves efficient segmentation
by selectively sampling multi-resolution patches centered on objects.
In the domain of active vision,
EyeRobot~\cite{kerr2025eye} explores learning gaze policies
for robotic manipulation via reinforcement learning.
Gazeformer~\cite{mondal2023gazeformer} and EyeFormer~\cite{jiang2024eyeformer}
utilize transformers to predict scanpaths in a supervised manner.
Finally, Glance-or-Gaze (GoG)~\cite{bai2026glance}
introduces a "selective gaze" mechanism for Large Multimodal Models,
dynamically choosing to consume global context
or focus on high-value regions via complexity-adaptive reinforcement learning.

\section{Conclusion}

In this work, we presented a novel \emph{sequential-to-global} self-supervised learning framework that successfully adapts the spatial DINO objective to dynamic, foveal-inspired vision transformers. 
By combining a top-down foveal grid with a highly efficient integral-image extraction method, we completely decouple the encoder's computational complexity from the input image resolution. 
Our results demonstrate that this architecture achieves competitive representation quality on ImageNet-1K and fine-grained benchmarks, enabling large-scale pretraining for image-size agnostic foundational models.

\noindent \textbf{Limitations and Future Work.} 
A primary limitation of our current pipeline lies in the sequential training of the task head and the active gaze policy. 
Currently, we train the task head using random gazes, freeze it, and then use its classification confidence as the reward signal to train the policy via GRPO. 
This creates a \textit{chicken-and-egg} optimization problem: if a challenging task requires isolating a very small, specific region of the image, random gazes are unlikely to sample it frequently enough to train a robust task head. 
Consequently, the weak frozen head provides an inadequate reward signal to teach the policy to find that exact region. 

\section*{Acknowledgements}
This research was partially funded by the dAIedge project
(HORIZON-CL4-2022-HUMAN-02-02, Grant Agreement
Number: 101120726) and the Ministry of Education and
Science of Bulgaria
(support for INSAIT, part of the Bulgarian National Roadmap for Research Infrastructure).
It was also supported with computational resources provided
by Google Cloud Platform (GCP).

%
%
\bibliographystyle{splncs04}
\bibliography{main}

\clearpage
\appendix
\section{Resolution-Agnostic Patch Extraction via Integral Images}

While the top-down patch extraction proposed in~\cite{prisadnikov2025vision}
achieves computational independence from image size in the model architecture, 
the initial tokenization process remains a bottleneck.
Standard patch extraction methods
typically exhibit linear complexity $\mathcal{O}(H_c \times W_c)$
with respect to the pixel count of the source crop,
leading to increased latency when processing high-resolution inputs.
Especially during training where we extract hundrerds of patches per image at each step.
To address this,
we introduce an efficient,
resolution-agnostic extraction mechanism based on \emph{Integral Images}~\cite{ViolaJ01},
originally introduced as Summed-Area Tables (SAT) by Crow~\cite{Crow1984SummedareaTF}.

Our approach decouples the computational cost of patch extraction from the source resolution.
By pre-computing the integral image representation,
we can extract and resize arbitrary rectangular regions into fixed-size tokens of dimension $P \times P$
(e.g., $16\times16$) with constant time complexity $\mathcal{O}(1)$ per target pixel,
or $\mathcal{O}(P^2)$ for the entire patch.

Let $I \in \mathbb{R}^{H \times W}$ denote the input image.
The integral image $S$ is defined such that $S(x, y)$
contains the sum of all pixels above and to the left of $(x, y)$:
\begin{equation}
    S(x, y) = \sum_{i \le x, j \le y} I(i, j)
\end{equation}
To extract a patch,
we define a grid of $P \times P$ target cells over the continuous crop region.
For a target pixel $v_{i,j}$,
let its corresponding source region $R$ be bounded
vertically by $[y_t, y_b]$ and horizontally by $[x_l, x_r]$.
The value of $v_{i,j}$ is computed via \emph{box filtering}:
\begin{equation}
    \label{eq:box_filter}
    v_{i,j} = \frac{1}{\text{Area}(R)} \left( S(x_r, y_b) + S(x_l, y_t) - S(x_l, y_b) - S(x_r, y_t) \right)
\end{equation}
This formulation allows for simultaneous cropping and resizing in a single operation.
To maximize computational throughput,
we employ a nearest-neighbor approximation
when mapping continuous crop coordinates to the discrete integral image grid,
rather than performing sub-pixel bilinear interpolation.
While this introduces minor quantization noise for non-aligned crops,
we observe that the resulting features maintain sufficient fidelity
for the downstream transformer while reducing overhead.
\section{Foveal Grid Formulation}

As shown in the paper,
extracting a grid of patches in a top-down manner
is an important part of our self-supervised pretraining --
either for extracting foveal grid of patches,
or extracting a ViT grid of patches.
Crucially, the grid of patches is extracted in a top-down manner.
The size of each patch and consequently of the whole grid
is defined through a zoom level $z$ which determines the size as a function of the image size.

%

\noindent \textbf{Grid Formulation.}
Let $(x_c, y_c)$ denote the center of gaze coordinates.
We define a focused grid of $G \times G$ number of patches (e.g. $4\times4$)
where each individual patch has the same fixed zoom level $z$.
The side length of a single patch in the grid, denoted as $S_{patch}$, is derived top-down from the image dimensions $(H, W)$:
\begin{equation}
    S_{patch} = \frac{\min(H, W)}{2^z}.
\end{equation}
The grid covers a total region of size $(G \cdot S_{patch}) \times (G \cdot S_{patch})$
centered at $(x_c, y_c)$.
The center coordinates $(x_{u,v}, y_{u,v})$ for a patch at grid index $(u, v)$,
where $u, v \in \{0, \dots, G-1\}$, are computed as:
\begin{equation}
    \begin{split}
        x_{u,v} &= x_c + \left( u - \frac{G-1}{2} \right) \cdot S_{patch} \\
        y_{u,v} &= y_c + \left( v - \frac{G-1}{2} \right) \cdot S_{patch}
    \end{split}
\end{equation}
This formulation centers the grid perfectly around the gaze point.
Using the integral image method,
we can extract and resize all $G^2$ patches in parallel.
This results in a composite context
consisting of the logarithmic multi-zoom patches (peripheral vision)
and the dense focused grid (foveal vision),
while maintaining a fixed input token length.

\noindent \textbf{Simulating random resized crops.}
Importantly, we can use the foveal grid approach
to simulate random resized crops as used in DINO's~\cite{caron2021emerging}
multi-crop setting~\cite{caron2020unsupervised}.
Consider a DINOv2 global crop.
It is a square random resized crop with scale above $0.32$,
i.e. the area of the crop is at least $32\%$ of the original image area.
Then the crop is resized to $224 \times 224$ and split into a $16 \times 16$ grid of patches,
i.e. the patch size $14\times14$ pixels.
In DINO~\cite{caron2021emerging} and DINOv3~\cite{simeoni2025dinov3}
the situation is the same, but with patch size of $16$,
and hence the crop is resized to $256 \times 256$.
If we represent the global crop as a foveal grid
we can simulate both the random resized crop and the patch extraction in a single step.
We set $G$ to $16$, then we choose zoom level range,
such that the total grid area has scale between $0.32$ and $1$.
Finally, we select the gaze center randomly so that the grid does not go beyond the image area.

\noindent \textbf{Selecting the zoom range.}
Without loss of generallity,
let us assume that the width and height of the image are $1$.
Then, given a zoom level $z$ the size of a single patch is $S_{patch} = 1/2^z$.
The total area of the crop is $G^2/2^{2z}$.
In order to satisfy that the area of the crop is at least $0.32$ we need
\begin{equation}
\begin{split}
    \frac{G^2}{2^{2z}} &\geq 0.32 \\
    2^{2z} &\leq \frac{G^2}{0.32} \\
    2z &\leq \log_2{G^2} - \log_2{0.32} \\
    z &\leq \log_2G - \frac{log_2{0.32}}{2}.\\
\end{split}
\end{equation}
Similarly, we can constrain the zoom level from above such that the scale is at most $1$.
\section{Pseudo-code for Self-supervised Pretraining}

Here we present a pseudo-code for the self-supervised training step
for our \emph{sequential-to-global} framework in Alg.~\ref{alg:training}.
Note that this is a heavily simplified version.
A lot of details would be challenging to be included in an easy to digest pseudocode.
Notably, although not included here we use KoLeo Loss~\cite{sablayrolles2018spreading}
similarly to DINOv2~\cite{oquab2023dinov2}.

\algblockdefx[With]{With}{EndWith}[1]{\textbf{with} #1 \textbf{do}}{\textbf{end with}}

\begin{algorithm}[h!]
\caption{Self-Supervised Pretraining of Foveal Active Vision Transformer}
\label{alg:training}
\begin{algorithmic}[1]
\Require Input images $X$, Student $\theta_s$, Teacher $\theta_t$, Steps $K$, Temp $\tau_s, \tau_t$
\Require Learned state embedding $\mathbf{e}_{state}$

\Statex \textsl{\# 1. Extract standard static crops and foveal multi-zoom sequences}
\State $V_g^s, V_g^t, V_l^s \gets \text{ExtractStaticCrops}(X)$
\Comment{Global and local ViT crops}
\State $L^s, L^t \gets \text{SampleGazeLocations}(X, K)$
\State $V_{mz}^s \gets \text{ExtractFoveal}(X, L^s)$
\Comment{Shape: $B \times K \times M \times P \times P \times 3$}
\State $V_{mz}^t \gets \text{ExtractFoveal}(X, L^t)$
\Comment{$M$: Total patches per location}

\Statex \textsl{\# 2. Teacher forward pass (No gradients)}
\With{no gradients}
    \State $Z_g^t \gets \theta_t(V_g^t, \mathbf{0} + \mathbf{e}_{state})$
    \Comment{Static crops omit state tracking}
    \State $s_0^t \gets \mathbf{0}$
    \Comment{The initial state is zeroes}
    \For{$k \in \{1, \dots, K\}$}
        \State $s_k^t, Z_{mz, k}^t \gets \theta_t(V_{mz, k}^t, s_{k-1}^t + \mathbf{e}_{state})$
    \EndFor
    \State $C^t \gets \text{SinkhornKnopp}(Z_g^t \cup \{Z_{mz, K}^t\}, \tau_t)$
    \Comment{Center global + final step}
\EndWith

\Statex \textsl{\# 3. Student forward pass}
\State $Z_g^s \gets \theta_s(V_g^s, \mathbf{e}_{state}), \quad Z_l^s \gets \theta_s(V_l^s, \mathbf{e}_{state})$
\State $\mathcal{L}_{static} \gets \frac{1}{8} \mathcal{H}(Z_g^s / \tau_s, C^t) + \frac{7}{8} \mathcal{H}(Z_l^s / \tau_s, C^t)$
\Comment{$\mathcal{H}$ is Cross-Entropy}

\Statex \textsl{\# 4. Student sequential pass (No BPTT)}
\State $s_0^s \gets \mathbf{0}, \quad \mathcal{L}_{mz} \gets 0$
\For{$k \in \{1, \dots, K\}$}
    \State $s_{k-1}^s \gets \text{stop\_gradient}(s_{k-1}^s)$
    \Comment{Prevent backprop through time}
    \State $s_k^s, Z_{mz, k}^s \gets \theta_s(V_{mz, k}^s, s_{k-1}^s + \mathbf{e}_{state})$
    \State $\mathcal{L}_{mz} \gets \mathcal{L}_{mz} + \mathcal{H}(Z_{mz, k}^s / \tau_s, C^t)$
\EndFor
\State $\mathcal{L} \gets \frac{1}{2} \left( \mathcal{L}_{static} + \frac{1}{K} \mathcal{L}_{mz} \right)$

\Statex \textsl{\# 5. Optimization}
\State $\theta_s \gets \theta_s - \eta \nabla_{\theta_s} \mathcal{L}$
\Comment{Update student via optimizer}
\State $\theta_t \gets \lambda \theta_t + (1 - \lambda)\theta_s$
\Comment{Exponential moving average ($\lambda$)}
\end{algorithmic}
\end{algorithm}
\section{Positional Embeddings for Multi-zoom Patches}

An important implementation detail not covered by the main paper
is handling positional embeddings for multi-zoom patches.
Unlike ViT patches we have patches of varying sizes (zoom levels).
This adds a necessity to include a third dimension when computing the positional embeddings.
Especially with the multi-zoom patches,
if the positional embeddings are function of the $x$ and $y$ coordinates
of the patch's center,
then all the multi-zoom patches will have the same positional embedding.
This is why in our work the positional embeddings are a function of $(x, y, z)$ --
the coordinates of the center of the patch and the zoom level $z$.

Unlike, standard ViT methods our positional embeddings are learned
through a shallow multi-layer perceptron (MLP)
which takes $(x, y, z)$ as three-dimensional input.

Note that this approach introduces a major change in
how random crops are represented.
Take an example of a random resized crop when training a standard ViT model.
The positional embeddings are such that the model perceives the crop
as if it is the entire image.
With our top-down approach and positional embeddings
the model implicitly "knows" that the random crop is part of a bigger image.
\section{Ablation Study Implementation Details}

For the ablation studies in the paper we use a lighter setting
in order to speed up the process of training.
We trained the model for 100 epochs,
while all scheduled hyperparameters like
learning rate, weight decay, teacher momentum, teacher temperature
are scheduled as if the training is 300 epochs long.
We also lowered the number of patches in the grid of ViT patches during training.
The global crops are still crops with scale above $0.32$,
but they are split into a $12 \times 12$ grid of patches.
Similarly the local crops are split into a $5 \times 5$ grid of patches.
The rest of the hyper-parameters are the same as the ones used for training the main model.

\end{document}